\pdfoutput=1
\documentclass[11pt]{article}

\usepackage{EMNLP2023}

\usepackage{times}
\usepackage{latexsym}

\usepackage[T1]{fontenc}
\usepackage[utf8]{inputenc}

\usepackage{microtype}

\usepackage{inconsolata}

\usepackage{amsmath}
\usepackage{makecell}
\usepackage{adjustbox}
\usepackage{multicol,multirow}
\usepackage{subfigure}
\usepackage{tabularx}
\title{MCC-KD: Multi-CoT Consistent Knowledge Distillation}

\author{Hongzhan Chen\textsuperscript{1}, Siyue Wu\textsuperscript{1}, Xiaojun Quan\textsuperscript{1}\thanks{\; Corresponding author.}, Rui Wang, Ming Yan\textsuperscript{2} and Ji Zhang\textsuperscript{2} \\
  \textsuperscript{1}School of Computer Science and Engineering, Sun Yat-sen University, China \\
  \textsuperscript{2}Alibaba Group, China \\
  \textsuperscript{1}\texttt{\{chenhzh59, wusy39\}@mail2.sysu.edu.cn, quanxj3@mail.sysu.edu.cn}\\
  \texttt{mars.wang@nyonic.ai} \\
  \textsuperscript{2}\texttt{\{ym119608, zj122146\}@alibaba-inc.com}} 

\begin{document}
\maketitle
\begin{abstract}

% Large language models (LLMs) have demonstrated impressive capability in complex reasoning via chain-of-thought (CoT) prompting. Transferring such reasoning ability from LLMs to smaller language models has attracted interest lately. However, transferring such reasoning ability could be challenging. To this end, we propose Multi-CoT Consistency Knowledge Distillation (MCC-KD), a novel approach of enforcing consistency among diverse rationales to efficiently distill reasoning capability. Comprehensive experiments are conducted with LLaMA and Flan-T5 on both the mathematical and commonsense reasoning benchmarks. The empirical results show: First, MCC-KD significantly enhances reasoning ability in small models. For example, the accuracy of the FlanT5-XXL on the GSM8K dataset is improved from 27.1\% to 33.99\%. Second, MCC-KD improves the generalization ability of small models. For instance, the accuracy of LLaMA-7B on the out-of-distribution (OOD) dataset SVAMP improves from 38.6\% to 44.0\%.

Large language models (LLMs) have showcased remarkable capabilities in complex reasoning through chain of thought (CoT) prompting.~Recently, there has been a growing interest in transferring these reasoning abilities from LLMs to smaller models.~However, achieving both the diversity and consistency in rationales presents a challenge.~In this paper, we focus on enhancing these two aspects and propose Multi-CoT Consistent Knowledge Distillation (MCC-KD) to efficiently distill the reasoning capabilities. In MCC-KD, we generate multiple rationales for each question and enforce consistency among the corresponding predictions by minimizing the bidirectional KL-divergence between the answer distributions.~We investigate the effectiveness of MCC-KD with different model architectures (LLaMA/FlanT5) and various model scales (3B/7B/11B/13B) on both mathematical reasoning and commonsense reasoning benchmarks. The empirical results not only confirm MCC-KD's superior performance on in-distribution datasets but also highlight its robust generalization ability on out-of-distribution datasets. Our code is available at \url{https://github.com/homzer/MCC-KD}.

%The empirical results demonstrate that MCC-KD achieves superior performance on in-distribution datasets and exhibits a strong generalization ability on out-of-distribution datasets.

%. Firstly, MCC-KD achieves better performance on in-distribution datasets. Secondly, MCC-KD improves the generalization ability of small models on out-of-distribution datasets.

% Specifically, for each question, MCC-KD minimizes the bidirectional KL-divergence between answer's distributions of diverse rationales from diverging to far. 

% Comprehensive experiments are conducted with LLaMA and Flan-T5 on both the mathematical and commonsense reasoning benchmarks. The empirical results show that: First, MCC-KD achieves better performance compared to current baseline methods. Second, MCC-KD improves the generalization ability of small models.

\end{abstract}

\section{Introduction}

Recently, large language models (LLMs) such as ChatGPT have exhibited impressive emergent capabilities, showcasing their competence in various tasks, including those demanding complex reasoning. While directly providing answers without generating intermediate steps may lead to errors and limited interpretability, chain of thought (CoT) \citep{chain-of-thought} prompting enables LLMs to break down reasoning tasks into a series of intermediate steps, guiding the model to generate the subsequent steps before arriving at the final answer. The effectiveness of CoT prompting has been demonstrated on diverse reasoning tasks \cite{kojima2022large}.

% Despite the outstanding performance of LLMs in reasoning abilities, they require a substantial amount of computational resources both during training and inference. Unfortunately, recent studies \citep{chain-of-thought,teaching-small2022,fu2023specializing} show that this emergent reasoning capability only appear in language models with over 100 billion parameters, such as PaLM-540B \citep{palm} or GPT-3 175B \citep{gpt-3}. The massive parameter size demands significant computational power and storage space, hindering deployment on resource-limited platforms. Another approach to utilize LLMs is through API calls. However, API calls are facing with three challenges: (1) API calls are limited by network stability; (2) API calls are challenging to localize and customize models; (3) The privacy of data prohibits direct transmission for remote fine-tuning.

Despite the effectiveness of CoT prompting, recent studies \citep{chain-of-thought,teaching-small2022,fu2023specializing} have shown that these reasoning capabilities only manifest in language models with over 100 billion parameters, such as PaLM (540B) \citep{chowdhery2022palm} and GPT-3 (175B) \citep{gpt-3}. These LLMs with massive parameter sizes require significant computational resources during both training and inference, which restrict their deployment on resource-limited platforms. While LLMs could be accessed through API calls, there are still several challenges to overcome, including network instability, difficulty in customizing the models, and privacy concerns.

%Despite the power of the CoT prompting, recent studies \citep{chain-of-thought,teaching-small2022,fu2023specializing} show that this emergent reasoning capability only appears in language models with over 100 billion parameters, such as PaLM-540B \citep{chowdhery2022palm} and GPT-3 175B \citep{gpt-3}. With such massive parameter sizes, these LLMs require a substantial amount of computational resources during both training and inference, hindering the deployment on resource-limited platforms. Even the LLMs can be utilized through API calls, there are still several obstacles to overcome: the network instability, difficult localization or customization of the models, and the privacy of the data prohibiting direct transmission for remote fine-tuning. % modified by wsy

\begin{figure}
    \centering
    \begin{adjustbox}{width=0.48\textwidth}
        \includegraphics{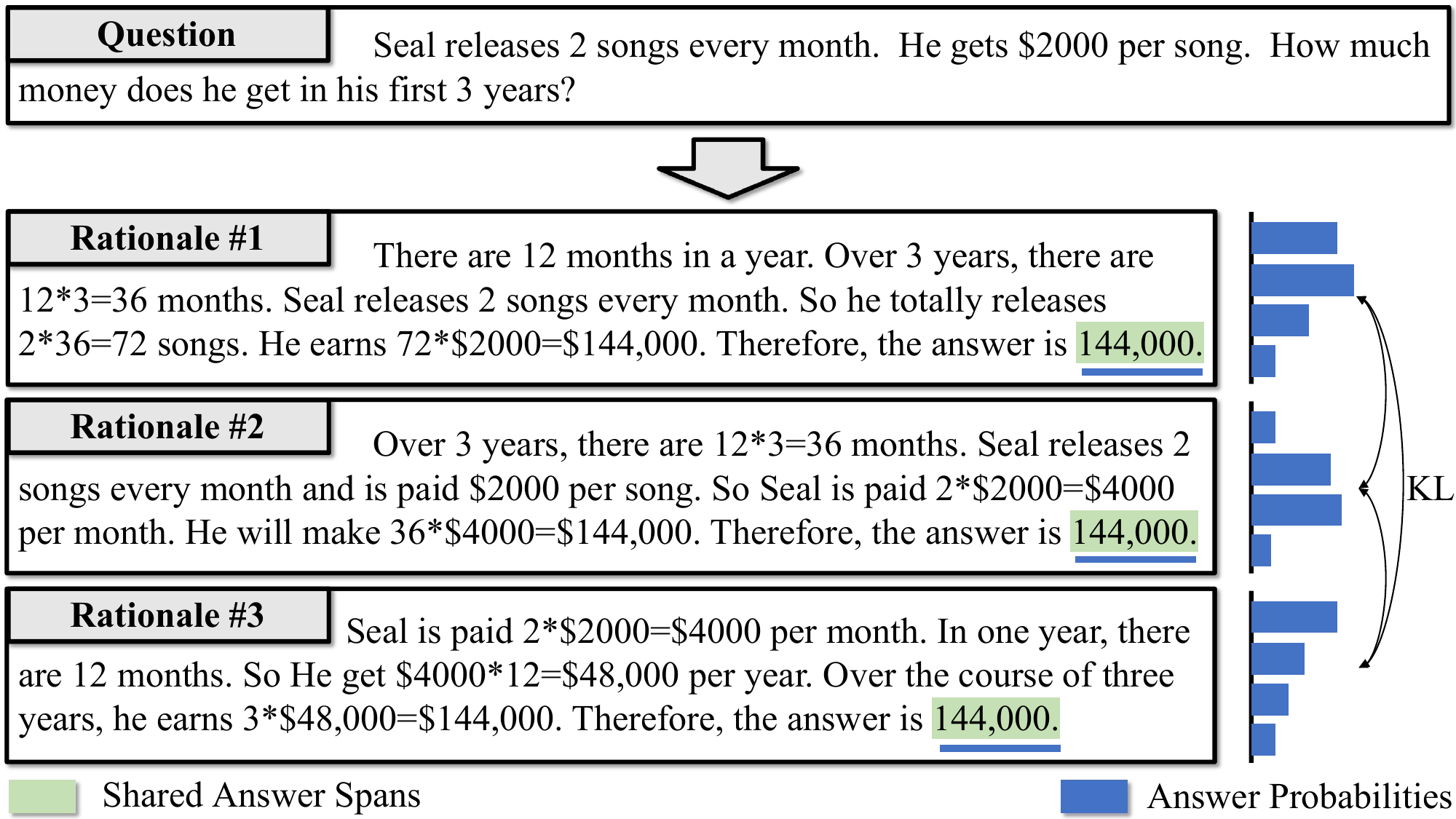}
    \end{adjustbox}
    \caption{An example question from the GSM8K dataset \citep{cobbe2021training} and the rationales generated by GPT-3.5-Turbo. Colored texts indicate the probabilities of the shared answer across different rationales.}
    \label{fig:example}
    \vspace{-3mm}
\end{figure}

% Therefore, an alternative solution is to deploy small language models such as LLaMA \citep{llama} or Flan-T5 \citep{flan-t5} (typically less than 13 billion parameters). An intuitive approach is to transfer the reasoning abilities of LLMs to smaller models via knowledge distillation (KD) \citep{knowledge-distilling}. Traditional knowledge distillation methods require the teacher model to provide output logits or hidden layer features, using KL divergence and MSE loss to guide the student model to mimic the teacher's behavior. However, due to the unavailability of weight parameters and direct access to intermediate features in current LLMs, traditional knowledge distillation methods do not work. As current LLMs are capable to generate high-quality reasoning data, one potential way is to distill the rationales generated by LLMs, transferring the reasoning abilities from the large teacher model to the smaller student models.

Therefore, an alternative approach is to deploy smaller language models such as LLaMA-7B/13B \citep{llama} and FlanT5-XL/XXL \citep{chung2022scaling}, which have fewer than 13 billion parameters. Through knowledge distillation (KD) \citep{knowledge-distilling}, the reasoning capabilities can be transferred from LLMs to these smaller models. However, traditional KD techniques require the teacher model to provide output logits or hidden layer features, which cannot be readily applied to LLMs due to the limited accessibility of their internals. One potential solution is to leverage rationales generated by LLMs to train smaller models, thereby acquiring their reasoning abilities \citep{ho2022reasoning-teacher,teaching-small2022,fu2023specializing}.

%Therefore, an alternative solution is to deploy smaller language models such as LLaMA-7B/13B \citep{llama} or FlanT5-XL/XXL \citep{chung2022scaling} (typically less than 13 billion parameters), to which the reasoning abilities of LLMs are transferred via knowledge distillation (KD) \citep{knowledge-distilling}. Traditional knowledge distillation methods require the teacher model to provide output logits or hidden layer features, using the KL-divergence or mean squared error (MSE) loss to guide the student model to mimic the teacher's behavior. However, due to the unavailability of the internal parameters of the current LLMs, traditional knowledge distillation methods do not work. Since the current LLMs are capable of generating high-quality reasoning steps, one potential way is to distill the rationales generated by LLMs, transferring the reasoning abilities from the large teacher models to the smaller student models \citep{ho2022reasoning-teacher,teaching-small2022,fu2023specializing}. % modified by wsy

% The idea of effectively leveraging the similarities and differences among distinct rationales has not been fully considered. 
However, these methods for rationale distillation also face several challenges. Firstly, the limited diversity in reasoning paths may lead to a dilemma where the student model simply mimics the superficial style of the teacher model's outputs \citep{false-promise} or overfits the training data, resulting in limited generalization capabilities. Secondly, despite the existence of multiple rationales leading to the same answer for each given question (as depicted in Figure~\ref{fig:example}), these methods neglect the consistency among different rationales in reaching the predicted answer when training the student model. Such oversights can undermine the stability of student models during training and impair their generalization capabilities.

% During the reasoning process, the model may generate rationales that lie outside the expected distribution due to randomness, even though they could still be correct. This can lead to different incorrect final answers.
% The common parts among different solutions may reveal the key steps toward the answer, while the distinct parts may help student generalize better. 
% (2) Using a filtering strategy based on Jaccard similarity \citep{jaccard} to acquire highly diverse rationales obtained from stochastic sampling.

To address these challenges, we propose Multi-CoT Consistent Knowledge Distillation (MCC-KD), a novel solution that incorporates two pivotal characteristics. Firstly, this approach leverages multiple diverse rationales for each given question and aims to improve their consistency in predicting the answer. This improvement is expected to enhance the stability and generalizability of the student models. Secondly, we introduce a similarity-based method to facilitate the selection of diverse rationales. MCC-KD draws inspiration from real-world teaching scenarios, where presenting multiple distinct solutions to one problem benefits the student's learning process. With these inherent advantages, MCC-KD enables the smaller models to acquire reasoning capabilities from larger models through effective knowledge distillation.

We conduct extensive experiments with LLaMA \citep{llama} and FlanT5 \citep{chung2022scaling} on both mathematical reasoning and commonsense reasoning benchmarks. The empirical results demonstrate the effectiveness and superiority of MCC-KD over previous CoT-based knowledge distillation methods. For example, MCC-KD achieves an accuracy improvement from point 38.01 to 41.58 on the GSM8K \citep{cobbe2021training} dataset with LLaMA-7B. Moreover, the generalization experiments reveal that MCC-KD achieves a substantial accuracy improvement, raising the performance from point 47.69 to 49.52 on the out-of-distribution dataset ASDiv \citep{miao-etal-2020-diverse} using FlanT5-XXL. These findings provide compelling evidence of the effectiveness and robustness of MCC-KD.

%We share three main findings from our experimental results: 
%First, with the constraint of multi-CoT consistency, MCC-KD effectively enhances the generalization ability of distilled models. 
%First, it is important not only to pursue a large quantity of teacher-generated rationales but also to consider the diversity of the rationales. This means that when obtaining more distilled reasoning data, diversity controlling is necessary. Second, for large datasets, it is not necessary to ensure the 100\% correctness of the teacher generated data, while a correctness rate ranging from 90\% to 95\% is still effective. 
% Third, generalized distillation performs almost as well as models specifically distilled on particular tasks. 
%Third, when conducting generalized distillation across multiple tasks, the model demonstrates a performance on a specific task that is comparable to the performance obtained from specialized distillation solely focused on that task.
% When conducting distillation across multiple tasks, the model exhibits a performance on a specific task that is on par with the results achieved through specialized distillation dedicated to that task.

\section{Related Work}

In this section, we briefly review the related work on chain of thought and knowledge distillation.

% Recently, there have been efforts to directly enable LLMs (Language Models) to generate rationales on their own for accomplishing complex tasks. The approach combines the Chain of Thought technique with in-context learning and few-shot prompting to guide LLMs in generating CoT (Chain of Thought) or rationales, thereby enhancing their ability to solve intricate reasoning tasks. Zero-shot CoT, on the other hand, leverages zero-shot prompting to guide LLMs, revealing their inherent capability to generate CoT or rationales without the need for manually-written contextual prompts.

\subsection{Chain of Thought}

The idea of using natural language rationales to solve mathematical problems through a series of intermediate steps is first pioneered by \citet{ling-etal-2017-program}.
Then it has been further shown that natural language rationales or intermediate steps can improve language models' performance \citep{yao2021refining,hase-bansal-2022-models} and robustness \citep{chen-etal-2022-rationalization} on various reasoning tasks. 
% Then it has been further shown that natural language intermediate steps can improve language models' performance \citep{zaidan-etal-2007-using,yao2021refining,hase-bansal-2022-models,gu2021dream} and robustness \citep{chen-etal-2022-rationalization} across various tasks. 
% By using natural language prompts \citep{gpt-3} as tasks descriptions \citep{t5,wei2021finetuned,sanh2022multitask}, LLMs are able to accomplish tasks based on human instructions \citep{instructgpt}. 
Following this idea, chain of thought \citep{chain-of-thought} prompting enables LLMs to generate CoTs or rationales themselves using in-context learning \citep{min2022rethinking} and few-shot prompting \citep{gpt-3}, thereby enhancing the models' capabilities to solve complex reasoning tasks.~\citet{self-consistency} introduce a multi-round voting mechanism to further improve the CoT prompting. On the other hand, \citet{kojima2022large} propose zero-shot-CoT prompting that leverages zero-shot prompting to guide LLMs, revealing their capabilities to generate CoTs or rationales without the need for manually-written contextual prompts.
% Chain of thought \citep{chain-of-thought,kojima2022large} prompting further enables LLMs to generate rationales, thereby enhancing the model's capability to solve complex reasoning tasks. 
However, \citet{hoffmann2022training} and \citet{chowdhery2022palm} unveil that CoT prompting requires the model's parameters to reach a certain scale to be effective.

% Chain-of-thought \citep{chain-of-thought,kojima2022large,self-consistency} prompting enables LLMs to generate reasoning steps, thereby enhancing the model's capability to solve complex reasoning task. CoT prompting is a model-agnostic approach that dose not require additional fine-tuning tasks, which improves performance over zero-shot learning and few-shot learning. However, \citet{hoffmann2022training} and \citet{chowdhery2022palm} indicate that the reasoning ability requires the models parameters to reach a certain scale (over 100 billion). Current LLMs have demonstrated the capability to generate high-quality rationales, and our work focuses on distilling this reasoning ability from large models to smaller ones.

% \subsection{Learning from rationales}

\begin{figure*}
    \centering
    \begin{adjustbox}{width=0.95\textwidth}
        \includegraphics{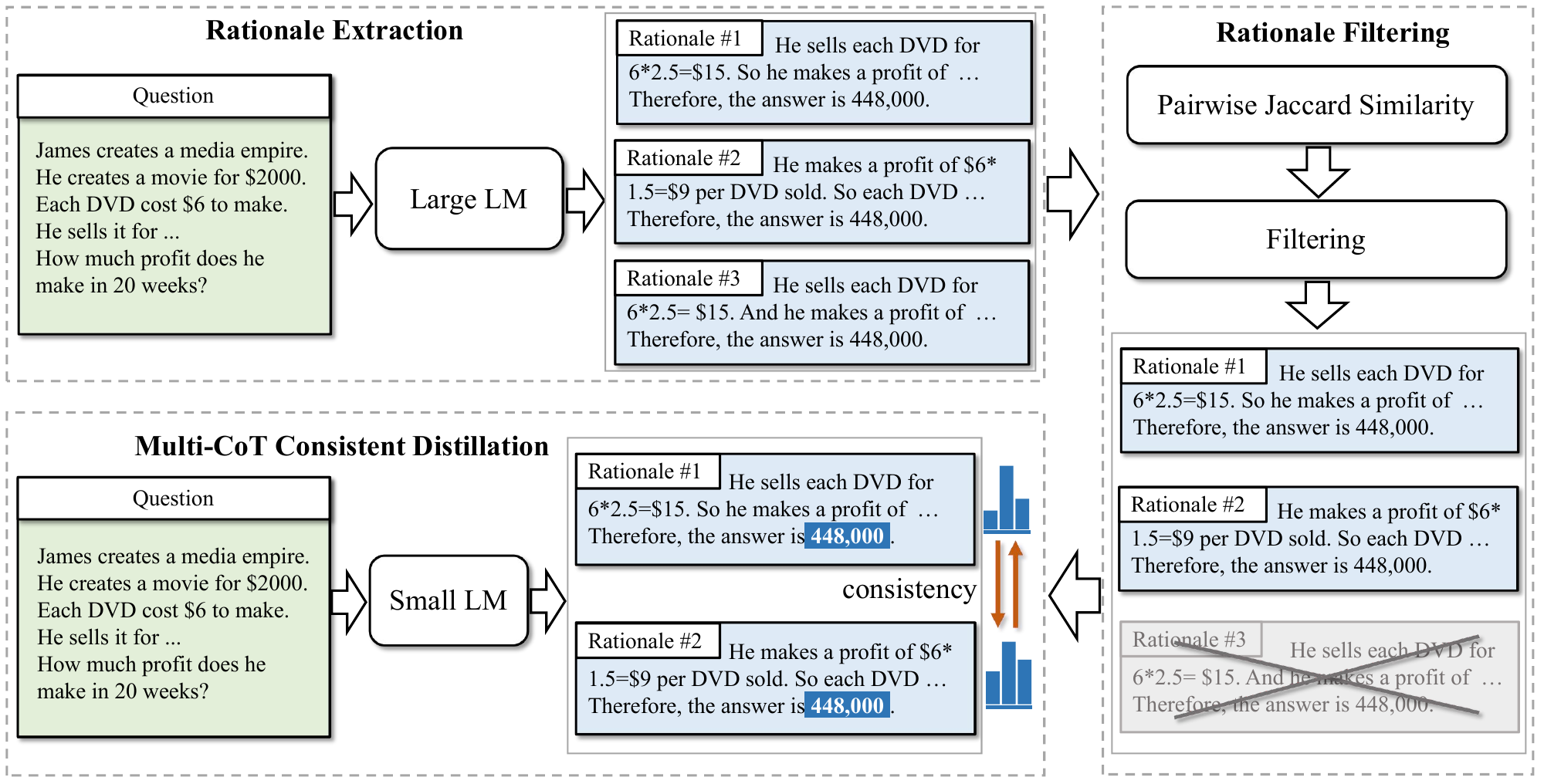}
    \end{adjustbox}
    \caption{\label{fig:method}Overview of the MCC-KD framework. Firstly, it leverages a LLM to generate multiple rationales. Subsequently, a filtering process is employed to preserve highly diversified rationales. Then the student model undergoes training by ensuring the consistency of the final answer predictions across various rationales.}
\end{figure*}

% In addition to using CoT prompting to stimulate the reasoning ability of LLMs, another way is to directly fine-tune the model via CoT data. \citet{large-self-improve} studies how LLMs can enhance reasoning ability by fine-tuning on the self-generated rationales. However, their work is limited to large models. \citet{explanations-form-llm} investigates the use of rationales generated by LLMs for direct fine-tuning of small models. However, the quantity and diversity of rationales should be considered when fine-tuning small models, or the small model may simply mimic the LLMs' style or overfit \citep{false-promise}. Our work focuses on both the quantity and diversity of rationales generated by LLMs.

% Aside from the CoT prompting technique, directly training with rationales is another approach. \citet{large-self-improve} shows that LLMs can use their generated rationales to improve themselves. In their work they also explore the teacher-student architecture. \citet{explanations-form-llm} fine-tuning a smaller model via LLMs generated rationales. However, recent works \citep{lima,false-promise} show that without simply mimicking the style of LLMs generated speech, the generated rationales or data should be sufficient.  

\subsection{Knowledge Distillation}

Knowledge distillation (KD) \citep{knowledge-distilling} aims to train smaller models by distilling knowledge from larger models, reducing model size while preserving high performance and generalization abilities. However, existing methods, such as response-based KD \citep{knowledge-distilling, turc2019well}, feature-based KD \citep{sun-etal-2019-patient}, and relation-based KD \citep{park-etal-2021-distilling}, all require access to the internal parameters of the teacher model, which are often impractical for LLMs.

Considering that many LLMs are capable of generating high-quality rationales, an alternative approach to knowledge distillation is to leverage the rationales generated by LLMs as distillation training data. Motivated by this, previous works \citep{distilling-multi-step, distilling-step-by-step, ho2022reasoning-teacher, teaching-small2022} employ LLMs as teacher models to generate chain of thought data as rationales, using the data to transfer the reasoning capabilities into smaller student models. Furthermore, \citet{fu2023specializing} specialize the model's ability towards a target task using chain of thought distillation. \citet{lion-adversarial-distillation} explore a teacher-feedback mechanism relying on LLMs to generate rationales of challenging instructions, aiding student models to learn from difficult samples. \citet{pinto} introduce a pipeline consisting of a rationalizing module and a reasoning module, resembling the teacher-student architecture, but it still requires LLMs to generate rationales for smaller models during the inference. The key distinction between these previous works and ours lies in that we explore the consistency among diverse rationales when training the student model to improve its stability.

\section{Method}

We introduce Multi-CoT Consistent Knowledge Distillation (MCC-KD), an approach designed to enhance the generalization and robustness of smaller student models during knowledge distillation. In particular, MCC-KD enforces the consistency among diverse chain of thoughts (CoTs) generated by the teacher LLMs in three key steps. Firstly, given an input question, we utilize zero-shot CoT prompting \citep{kojima2022large} to obtain multiple CoTs as rationales from a LLM such as GPT-3.5. Secondly, we filter out rationales that exhibit limited diversity. Lastly, we impose constraints on the outputs from multiple rationales to facilitate their consistency. The overall framework of MCC-KD is illustrated in Figure~\ref{fig:method}.

\subsection{Rationale Extraction}
Following \citep{kojima2022large}, we prompt the teacher model to generate rationales for each question. Formally, let $x$ represent a question, $r$ denote a rationale generated by the teacher, and $a$ indicate the final answer predicted by the teacher from $x$ and $r$. A resulting training sample is constructed by concatenating $x$, $r$, and $a$ in the following format: \texttt{<$x$> <$r$> Therefore, the answer is <$a$>}.

% To ensure diversity and quantity of the rationales, we increase the sampling temperature $T$ (typically, we set $T=1.3$ in our experiments) as well as the number of sampling iterations. To ensure the correctness of the rationales, we filter based on the predicted answer $\hat{a}$ and the label. Notably, for non-multiple-choice tasks, the correctness of the answer is usually consistent with the correctness of the rationales (from our observation, 99\% of the cases). However, for multiple-choice tasks, it is challenging to guarantee the same level of certainty. We manually check for their correctness.

To ensure the abundance and diversity of the generated rationales, we employ a combination of techniques. Firstly, we increase the sampling temperature $\tau$ ($\tau=1.3$ in this study) along with the number of sampling iterations. This approach aids in generating a greater variety of rationales. Secondly, to ensure the accuracy of the generated samples, we conduct verification for both the answer and the rationale. For answer verification, we compare the predicted answer $a$ with the ground truth answer to confirm its correctness. For rationale verification, we find that the correctness of rationales typically aligns with the correctness of answers under most circumstances.
% For rationale verification, we manually check the correctness of the rationales.
% the correctness of the answer typically aligns with the correctness of the rationales in non-multiple-choice tasks. However, in multiple-choice tasks, the alignment is challenging due to the possibility of having multiple correct answers for each question. Therefore, we manually verify the rationales in such cases.

\subsection{Rationale Filtering} 
\label{sec:filtering}
The diversity of reasoning rationales plays a crucial role in transferring reasoning capabilities from teacher LLMs to student models \citep{false-promise}. However, based on our observations, the teacher model still tends to generate similar rationales even with different sampling temperatures (as shown in Table~\ref{tab:repetitiveness} in Appendix~\ref{sec:appendix-diversity}). In order to obtain more diverse rationales, we develop a filtering strategy based on N-gram. For each rationale, we convert it into a set of N-gram (specifically, 3-gram in this study) segments. Subsequently, we calculate the Jaccard similarity among these sets. To be more specific, considering that there are $M$ rationales $\{r_1, r_2, \dots, r_M\}$ extracted for the input question, each rationale $r_i$ is decomposed into a set $S_i$ of segments. We then compare each pair of segment sets using the Jaccard similarity score to identify the most similar rationales:
\begin{equation}
    (k, l)=\underset{1\leq i,j\leq M, i\neq j}{\arg\max}\frac{|{S}_i\cap{S}_j|}{|{S}_i\cup{S}_j|},
\end{equation}
where $k$ and $l$ represent the indices of the selected rationale pair, $r_k$ and $r_l$, respectively. 
Subsequently, we randomly retain one of the two rationales while discarding the other. This iterative process continues until we accumulate a predefined number, denoted as $K$ (set to 5 in our experiments), of rationales for the given question.

%We then randomly keep one of the two rationales and discard the other. This process is repeated until we obtain a predetermined number $K$ (set to 5 in our experiments) of final rationales for the question.

% For pairs with the maximum similarity, we merge them by keeping only one of the two. A threshold can be set, where we only merge pairs that have a similarity greater than the threshold. 

\subsection{Multi-CoT Consistent Distillation}

%When faced with a question, there may be multiple distinct solutions or rationales to arrive at the correct answer. The common parts of the answer are shared by different rationales. 

As previously discussed, LLMs typically generate multiple valid rationales for a given input question. This work is built on the assumption that ensuring consistency among the predicted answers is crucial when training the student model with these rationales. For the input question and the $K$ retained rationales ($r_1, r_2, ..., r_K$) after filtering, we ensure consistency in the predictions of the student model by minimizing the variations in the probabilities of the answers from these different rationales.

\noindent \textbf{Single-token answer}~ We first consider the scenario where the answer consists of a single token, where the prediction corresponds to a probability distribution over the vocabulary. For a given rationale $r_i$, let $\boldsymbol{p}$ represent the predicted distribution obtained for the answer, and for another rationale $r_j$, let $\boldsymbol{q}$ represent the predicted distribution. In order to ensure consistency between these two rationales, we apply bidirectional KL-divergence to their corresponding distributions as our training objective:
\begin{equation}
    \mathcal{L}_{kl}(\boldsymbol{p},\boldsymbol{q})=\sum^V_{i=1} (p_i\log{\frac{p_i}{q_i}}+q_i\log{\frac{q_i}{p_i}}),
    \label{eq:kl-1}
\end{equation}
where $V$ denotes the size of the vocabulary.

\noindent \textbf{Multi-token answer}~As for the answer consisting of $T$ tokens, each token having its own distribution, we define $\boldsymbol{P}=\{\boldsymbol{p}_1,\boldsymbol{p}_2,\dots,\boldsymbol{p}_T\}$ as the set of predicted distributions for the answer obtained through rationale $r_i$, where $\boldsymbol{p}_t$ represents the probability distribution of the $t$-th token in the answer. Similarly, we use $\boldsymbol{Q}=\{\boldsymbol{q}_1,\boldsymbol{q}_2,\dots,\boldsymbol{q}_T\}$ to represent the predicted distributions obtained through rationale $r_j$. To achieve multi-CoT consistency, we calculate the bidirectional KL-divergence for each token according to Equation~\ref{eq:kl-1} and take the average divergence to obtain the training objective:
\begin{equation}
    \mathcal{L}_{kl}(\boldsymbol{P},\boldsymbol{Q})=\frac{1}{T}\sum^T_{t=1}\mathcal{L}_{kl}(\boldsymbol{p}_t,\boldsymbol{q}_t).
\end{equation}
% KL equals KL1 when T=1, and for T>=2, KL remains the same as KL1.
% When $T=1$, let $\mathcal{L}_{kl}=\mathcal{L}^{single}_{kl}$, and for $T\geq2$, $\mathcal{L}_{kl}=\mathcal{L}^{multi}_{kl}$. 

\noindent \textbf{Pairwise rationale sampling} Since there are $K$ rationales available for each question, we randomly select two distinct ones from the set of rationales in each training epoch to compute the $\mathcal{L}_{kl}$ loss.

% we randomly select two distinct rationales out of $K$ available rationales, 

%\begin{equation}
%    \{r_i,r_j\}=\underset{i\neq j}%{\text{UniformSample}}(\{r_1,\dots,r_K\})
%\end{equation}

\subsection{Overall Objective}

% We combine the original cross-entropy loss between the output of student and the ground-truth label, and the KL-divergence loss term. The overall objective is defined as:

% The overall objective is define as the combination of the original cross-entropy loss between the outputs of the student model and the ground-truth labels, along with the KL-divergence 

The overall objective function is defined as a combination of the cross-entropy loss ($\mathcal{L}_{ce}$) in traditional causal language modeling, computed on rationale and answer tokens, and the multi-CoT consistent loss ($\mathcal{L}_{kl}$), which ensures consistency in the model's answer distribution. The objective function can be represented as follows:
\begin{equation}
\mathcal{L} = \mathcal{L}_{ce} + \alpha\mathcal{L}_{kl},
\end{equation}
where $\alpha$ is a hyperparameter used to adjust the strength of the KL-divergence constraint.

%This also differentiates us from other CoT-based knowledge distillation methods \citep{teaching-small2022,ho2022reasoning-teacher,distilling-step-by-step,fu2023specializing}.

% \subsection{Other Methods}
% MCC-KD is similar to \citet{wu2021r}, but we add constraints between distinct rationales instead of the sub-modules of dropout. 
%Different from other CoT-based knowledge distillation methods \citep{teaching-small2022,ho2022reasoning-teacher,distilling-step-by-step,fu2023specializing}, we focus more on the consistency among diverse rationales.

\section{Experimental Setup}

In this section, we present the datasets and backbone models utilized in our experiments.

\subsection{Datasets}

To evaluate our method, we adopt both mathematical reasoning and commonsense reasoning tasks, following \citet{ho2022reasoning-teacher} and \citet{fu2023specializing}. For in-distribution mathematical reasoning, we employ the benchmarks GSM8K~\citep{cobbe2021training}, SVAMP~\citep{patel-etal-2021-nlp}, and ASDiv~\citep{miao-etal-2020-diverse}. Additionally, we also employ out-of-distribution (OOD) mathematical reasoning benchmarks to assess the OOD generalization capability, including SingleEq~\citep{koncel-kedziorski-etal-2015-parsing}, AddSub~\citep{hosseini-etal-2014-learning}, and MultiArith~\citep{roy-roth-2015-solving} from the Math World Problem Repository~\citep{koncel-kedziorski-etal-2016-mawps}. In the realm of commonsense reasoning, we employ CommonsenseQA~\citep{talmor-etal-2019-commonsenseqa} as the in-distribution dataset. For OOD evaluations of commonsense reasoning, we utilize Date Understanding, Tracking Shuffled Objects from the BIG-bench \citep{big-bench}, Coin Flip from \citet{kojima2022large}, as well as the StrategyQA \citep{geva-etal-2021-aristotle} dataset. Further statistics of these datasets are provided in Appendix~\ref{sec:dataset}.

% Following~\cite{ho2022reasoning-teacher} we mainly choose two categories to evaluate our method: categorizing arithmetic reasoning and commonsense reasoning. We mainly focus on transferring arithmetic reasoning capabilities from LLMs to smaller models since arithmetic reasoning is much complicated. For arithmetic reasoning, we evaluate on SingleEq~\citep{singleEq}, AddSub~\citep{addsub} and MultiArith~\citep{multiarith} from the Math World Problem Repository~\citep{math-world-problems} as well as more recent datasets, GSM8K~\citep{gsm8k} and SVAMP~\citep{svamp}. For commonsense reasoning, we use CommonsenseQA~\citep{commonsenseQA}. We provide details on datasets used in Appendix[].

\subsection{Backbone Models}

We use GPT-3.5-Turbo as the teacher model and prompt it to generate chain of thought samples (rationales). Following the filtering process introduced in Section \ref{sec:filtering}, we retain $K$=5 rationales for each question across all our training datasets. As for the student models, we employ the instruction-tuned FlanT5-XL/XXL (3B/11B) \citep{chung2022scaling} and LLaMA-7B/13B \citep{llama}, which are initialized with pre-trained weights obtained from Hugging Face\footnote{\url{https://huggingface.co/models}}. For the purpose of accelerating training and conserving GPU memory, we apply LoRA \citep{lora} throughout all of our experiments. The model configurations are summarized in Table~\ref{tab:model-config}, and additional details regarding the settings can be found in Appendix~\ref{sec:model}.

\begin{table}[ht]
    \centering
    \renewcommand{\arraystretch}{1.2}
    \begin{adjustbox}{width=0.48\textwidth}
        \begin{tabular}{l|cccc}
         \hline
         Models & \makecell[c]{Sequence \\ Length} & \#GPUs & \makecell[c]{LoRA \\ Rank} & Precision \\
         \hline
         \hline
         FlanT5-XL & 196/384 & 4 & 64 & float32 \\
         FlanT5-XXL & 196/384 & 8 & 128 & float32 \\
         LLaMA-7B & 512 & 4 & 64 & float16 \\
         LLaMA-13B & 512 & 8 & 128 & float16 \\
         \hline
    \end{tabular}
    \end{adjustbox}
    \caption{Model configurations. For FlanT5 models, the encoder and decoder have input lengths of 196 and 384, respectively. For LLaMA models, the input length is 512. To optimize memory usage and accelerate training, we leverage LoRA \citep{lora} and employ varying data precision. All models are trained on multiple GPUs and adopt a greedy decoding strategy.}
    \label{tab:model-config}
\end{table}

\subsection{Baseline Methods}

In order to evaluate the effectiveness of MCC-KD, we conduct experiments and compare its performance with existing CoT-based distillation methods~\citep{ho2022reasoning-teacher, fu2023specializing, teaching-small2022}, which utilize LLMs as teacher models to generate rationales and distill their reasoning abilities directly into smaller student models. For a fair comparison, we also implement a baseline method called Vanilla KD on our training datasets. Vanilla KD is a CoT-based distillation method that does not incorporate diversity filtering and the multi-CoT consistency constraint.

\begin{table*}[t]
    \centering
    \vspace{-4mm}
    \renewcommand{\arraystretch}{1.2}
    \begin{adjustbox}{width=0.85\textwidth}
        \begin{tabular}{l|c|ccc|c}
        \hline
        \textbf{Models} & \textbf{\# Params} & \textbf{GSM8K} & \textbf{ASDiv} & \textbf{SVAMP} & \makecell[c]{\textbf{CommonsenseQA}} \\
        \hline
        \hline
        GPT-3.5-Turbo (teacher) & - & 73.98 & 79.64 & 75.14 & 74.35 \\
        \hline
        GPT-3-babbage~\citep{ho2022reasoning-teacher} & 1.3B & 4.70 & - & 8.00 & 43.08 \\
        GPT-3-curie~\citep{ho2022reasoning-teacher} & 6.7B & 6.75 & - & 12.67 & 56.76 \\
        T5-XXL~\citep{teaching-small2022} & 11B & 21.99 & 42.12 & - & - \\
        FlanT5-XL~\citep{fu2023specializing} & 3B & 22.4 & 28.4 & 23.8 & - \\
        FlanT5-XXL~\citep{fu2023specializing} & 11B & 27.1 & 37.6 & 35.6 & - \\
        FlanT5-XL (Vanilla KD) & 3B & 22.76 & 29.41 & 29.33 & 81.13 \\
        FlanT5-XXL (Vanilla KD) & 11B & 33.33 & 48.24 & 51.33 & 84.32 \\
        LLaMA-7B (Vanilla KD) & 7B & 38.01 & 64.01 & 62.67 & 75.10 \\
        LLaMA-13B (Vanilla KD) & 13B & 47.19 & 68.79 & 68.0 & 78.42 \\
        \hline
        FlanT5-XL (MCC-KD) & 3B & 24.28 & 31.35 & 30.0 & 82.88 \\
        FlanT5-XXL (MCC-KD) & 11B & 33.99 & 48.73 & 52.67 & \textbf{84.93} \\
        LLaMA-7B (MCC-KD) & 7B & 41.58 & 65.76 & 64.67 & 76.41 \\
        LLaMA-13B (MCC-KD) & 13B & \textbf{48.71} & \textbf{69.11} & \textbf{68.66} & 78.46 \\
        \hline
        \end{tabular}
    \end{adjustbox}
    \caption{\label{tbl:main-results}
    Overall test accuracy for arithmetic and commonsense reasoning tasks. The reported results are averaged over three runs using randomly selected seeds. Baseline results from other studies \cite{ho2022reasoning-teacher,teaching-small2022,fu2023specializing} are included, while the performance of GPT-3.5-Turbo is assessed through our own evaluation.}
    \vspace{-2mm}
\end{table*}

\section{Results and Analysis}

In this section, we present the main results, ablation studies, and additional experiments.

\subsection{Main Results}
The main results for mathematical and commonsense reasoning tasks are provided in Table~\ref{tbl:main-results}. The baseline results are obtained from their respective original papers \citep{ho2022reasoning-teacher,teaching-small2022,fu2023specializing}. We evaluate the performance of the teacher model through our own testing. It can be observed that MCC-KD outperforms current baseline methods in all mathematical reasoning tasks, namely GSM8K, ASDiv, and SVAMP, when compared with models of similar size. These results highlight significant improvements achieved by MCC-KD. 
The performance gap between the FlanT5 (Vanilla KD) that we implement and \citet{fu2023specializing} can be explained by the difference in teacher model selection. \citet{fu2023specializing} utilize code-davinci-002 as the teacher model, while we utilize GPT-3.5-turbo as the teacher model.
For the commonsense reasoning tasks, MCC-KD surpasses current baseline methods and even exceeds the performance of the teacher model on the CommonsenseQA dataset. This outcome clearly demonstrates the effectiveness of MCC-KD in addressing commonsense reasoning tasks. Notably, the distilled models are able to generate reasoning paths directly, eliminating the necessity for any CoT prompting throughout our experiments.

\subsection{Ablation Study}

This ablation study aims to examine the influence of components in MCC-KD. Results are averaged over three runs using randomly selected seeds.

\noindent \textbf{Multi-CoT consistency constraint}~To assess the impact of the multi-CoT consistency constraint, we perform ablation experiments on different variants of MCC-KD using LLaMA-7B models without the consistency constraint. As presented in Table~\ref{tab:ablation}, we observe a significant decrease in performance on both mathematical and commonsense reasoning tasks when the consistency constraint is removed.

\begin{table}[t]
    \centering
    \renewcommand{\arraystretch}{1.2}
    \begin{adjustbox}{width=0.48\textwidth}
        \begin{tabular}{l|ccc|c}
        \hline
         \textbf{Method} & \textbf{GSM8K} & \textbf{ASDiv} & \textbf{SVAMP} & \makecell[c]{\textbf{Common} \\ \textbf{SenseQA}} \\
        \hline
        \hline
        MCC-KD & 41.67 & 65.18 & 64.17 & 74.28 \\
        \ \ w/o $\mathcal{L}_{kl}$ & 40.45 & 64.22 & 63.28 & 73.20 \\
        \ \ w/o filtering & 39.55 & 63.90 & 62.96 & 73.49 \\
        \hline
    \end{tabular}
    \end{adjustbox}
    \caption{Results of ablation study of multi-CoT consistency and diversity filtering on development sets.}
    \label{tab:ablation}
    \vspace{-3mm}
\end{table}

\noindent \textbf{Rationale filtering} ~We then explore the effectiveness of rationale filtering in MCC-KD. In the experiment setting without rationale filtering, we randomly sample 5 rationales for each question, maintaining the same quantity as the experiment setting with rationale filtering. Additionally, we ensure that the correctness rate of the selected rationales remains consistent between both experiment settings. As demonstrated in Table~\ref{tab:ablation}, we observe a noticeable decline in performance after removing the rationale filtering process, highlighting the critical importance of rationale diversity.

\noindent \textbf{Student model architecture} ~Furthermore, we examine the effectiveness of MCC-KD on two distinct model architectures: FlanT5-XL for the encoder-decoder Transformer architecture \citep{attention-is-all-you-need} and LLaMA-7B for the decoder-only architecture. We compare MCC-KD with the vanilla knowledge distillation (KD) approach. As presented in Table~\ref{tbl:main-results}, MCC-KD consistently enhances performance in comparison to vanilla KD across various model architectures.

% \begin{table}[t]
%     \centering
%     \renewcommand{\arraystretch}{1.2}
%     \begin{adjustbox}{width=0.48\textwidth}
%         \begin{tabular}{l|ccc|c}
%             \hline
%             \textbf{Models \& Methods} & \textbf{GSM8K} & \textbf{ASDiv} & \textbf{SVAMP} & \makecell[c]{\textbf{Common} \\ \textbf{SenseQA}} \\
%             \hline
%             \hline
%             FlanT5-XL (Vanilla KD) & 22.76 & 29.41 & 29.33 & 81.13 \\
%             FlanT5-XL (MCC-KD) & 24.28 & 31.35 & 30.0 & \textbf{82.88} \\
%             \hline
%             LLaMA-7B (Vanilla KD) & 38.01 & 64.01 & 62.67 & 75.10 \\
%             LLaMA-7B (MCC-KD) & \textbf{41.58} & \textbf{65.76} & \textbf{64.67} & 76.41 \\
%             \hline
%         \end{tabular}
%     \end{adjustbox}
%     \caption{\label{tbl:ablation-vanilla}Test performances of MCC-KD and vanilla KD on various model architectures and datasets.}
% \end{table}

\begin{table*}[ht]
    \centering
    \vspace{-2mm}
    \renewcommand{\arraystretch}{1.2}
    \begin{adjustbox}{width=0.95\textwidth}
    \begin{tabular}{l|c|c|ccccc|c}
         \hline
         \textbf{Models} & \textbf{\# Params} & \textbf{GSM8K} & \textbf{ASDiv} & \textbf{SVAMP} & \textbf{MultiArith} & \textbf{SingleEq} & \textbf{AddSub} & \textbf{Avg} \\
         \hline
         \hline
         FlanT5-XL~\citep{fu2023specializing} & 3B & 22.4 & 28.4 & 23.8 & 42.3 & - & - & - \\
         FlanT5-XXL~\citep{fu2023specializing} & 11B & 27.1 & 37.6 & 35.6 & 63.0 & - & - & - \\
         FlanT5-XL (Vanilla KD) & 3B & 22.76 & 26.84 & 24.67 & 42.0 & 26.84 & 16.65 & 27.4 \\
         FlanT5-XXL (Vanilla KD) & 11B & 33.33 & 47.69 & 39.67 & 78.0 & 46.26 & 37.82 & 49.89 \\
         FlanT5-XL (MCC-KD) & 3B & 24.28 & 28.98 & 26.67 & 44.44 & 27.32 & 15.58 & 28.6 \\
         FlanT5-XXL (MCC-KD) & 11B & 33.99 & 49.52 & 38.67 & 77.78 & 47.06 & 39.50 & 50.51 \\
         \hline
         LLaMA-7B (Vanilla KD) & 7B & 38.01 & 56.37 & 39.3 & 84.44 & 52.94 & 43.69 & 55.35 \\
         LLaMA-13B (Vanilla KD) & 13B & 47.19 & 65.18 & 55.34 & 91.11 & 62.75 & 51.38 & 65.15 \\
         LLaMA-7B (MCC-KD) & 7B & 41.58 & 57.64 & 41.0 & 86.67 & 54.90 & 45.38 & 57.12 \\
         LLaMA-13B (MCC-KD) & 13B & 48.71 & \textbf{66.45} & \textbf{57.33} & \textbf{93.33} & \textbf{61.45} & \textbf{52.10} & \textbf{66.13} \\
         \hline
    \end{tabular}
    \end{adjustbox}
    \caption{\label{tbl:ood}Out-of-distribution performance of MCC-KD on mathematical reasoning. We train our models on the in-distribution dataset (GSM8K) and evaluate the best checkpoints on the out-of-distribution datasets (ASDiv, SVAMP, MultiArith, SingleEq, and AddSub).}
\end{table*}

\begin{table*}[ht]
    \centering
    \renewcommand{\arraystretch}{1.2}
    \begin{adjustbox}{width=0.95\textwidth}
    \begin{tabular}{l|c|c|cccc}
         \hline
         \textbf{Models} & \textbf{\# Params} & \makecell[c]{\textbf{Common} \\ \textbf{SenseQA}} & \textbf{StrategyQA} & \makecell[c]{\textbf{Date} \\ \textbf{Understanding}} & \makecell[c]{\textbf{Shuffled} \\ \textbf{Objects}} & \makecell[c]{\textbf{Coin} \\ \textbf{Filp}} \\
         \hline
         \hline
         FlanT5-XL (Vanilla KD) & 3B & 81.13 & 65.74 & 46.2 & 30.8 & 46.2 \\
         FlanT5-XL (MCC-KD) & 3B & 82.88 & \textbf{67.05} & \textbf{46.61} & 30.4 & 48.0 \\
         \hline
         LLaMA-7B (Vanilla KD) & 7B & 75.10 & 52.77 & 43.36 & 30.53 & \textbf{49.6} \\
         LLaMA-7B (MCC-KD) & 7B & 76.41 & 57.43 & 45.53 & \textbf{33.33} & \textbf{49.6} \\
         \hline
    \end{tabular}
    \end{adjustbox}
    \caption{\label{tbl:ood-commonsense}Out-of-distribution performance of MCC-KD on commonsense reasoning. We train our models on the in-distribution dataset (CommonsenseQA) and evaluate the best checkpoints on the out-of-distribution datasets (StrategyQA, Date Understanding, Tracking Shuffled Objects and Coin Filp).}
    \vspace{-8mm}
\end{table*}

\subsection{Out-of-Distribution Generalization}

In line with the work of \citet{fu2023specializing}, we explore the ability of MCC-KD to enhance the generalization capabilities of models. We apply MCC-KD to the in-distribution mathematical reasoning dataset (GSM8K) and select the optimal checkpoints for evaluation on out-of-distribution mathematical reasoning datasets (ASDiv, SVAMP, MultiArith, SingleEq, and AddSub). Similarly, we assess the generalization performance on commonsense reasoning tasks using both the in-distribution dataset (CommonsenseQA) and out-of-distribution datasets (StrategyQA, Date Understanding, Tracking Shuffled Objects, and Coin Flip). The results are presented in Table~\ref{tbl:ood} and Table~\ref{tbl:ood-commonsense}, respectively. In mathematical reasoning, MCC-KD demonstrates further improvements in models' generalization capabilities compared to the findings of \citet{fu2023specializing} as well as the vanilla KD approach. In commonsense reasoning, MCC-KD exhibits a consistent trend of enhancing generalization capabilities when compared to the vanilla KD approach.

\begin{figure}[ht]
    \centering
    \vspace{-0mm}
    \begin{adjustbox}{width=0.45\textwidth}
    \includegraphics{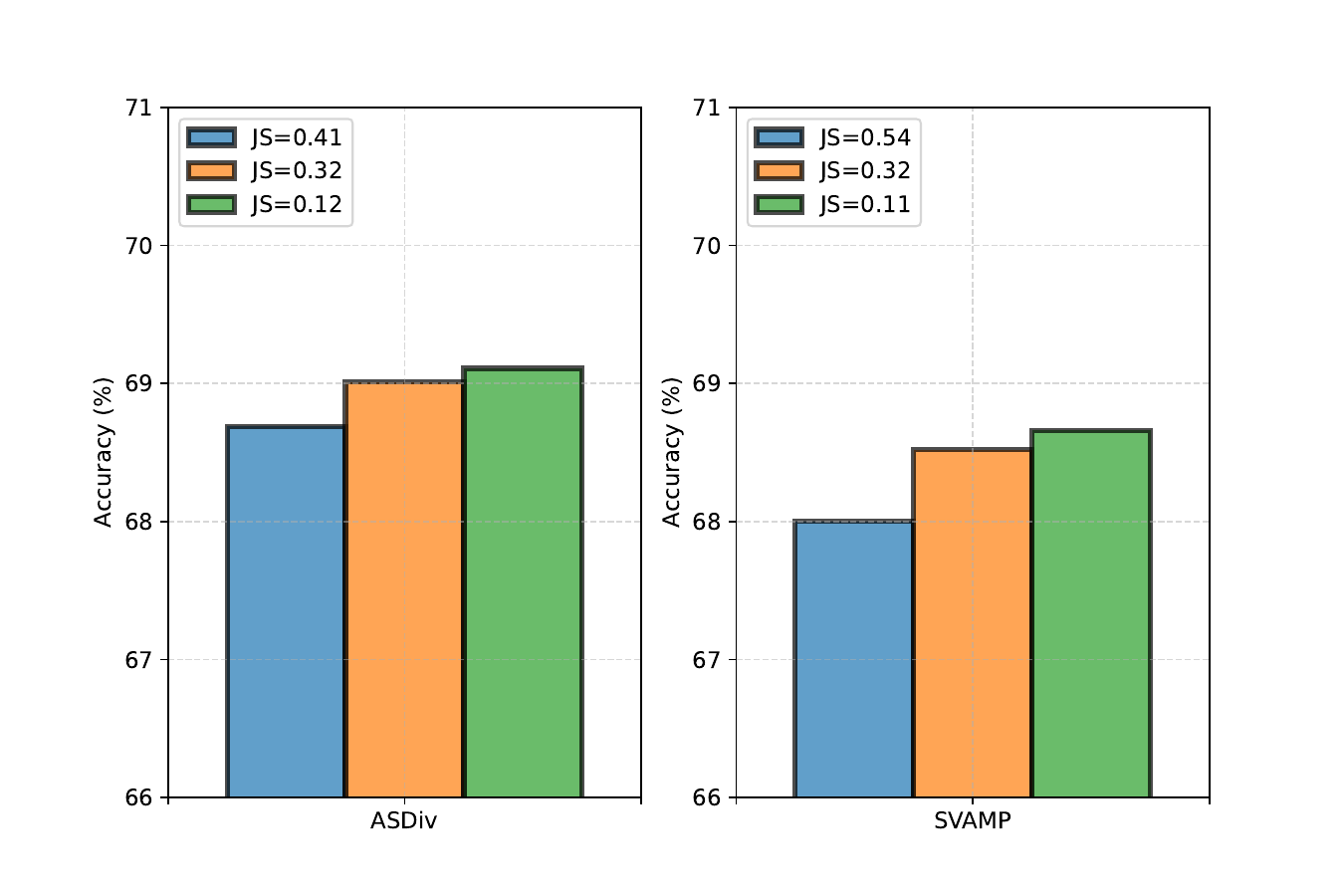}
    \end{adjustbox}
    \caption{The performance of MCC-KD on ASDiv and SVAMP development sets with different rationale diversities. JS stands for Jaccard similarity.}
    \label{fig:diversity}
    \vspace{-3mm}
\end{figure}

\subsection{Diversity of Rationales}
In this section, we delve into the significance of rationale diversity within the context of MCC-KD. We contend that diversity among the rationales generated by the teacher model is a crucial factor for effective reasoning learning by the student model. To measure the degree of diversity among rationales, we employ the Jaccard similarity metric, where a higher score indicates greater similarity and vice versa. By manipulating the diversity of training instances, we assess the efficacy of MCC-KD. In our experiments, we utilize the LLaMA-13B model as the student model and incorporate two rationales for each question during training. As illustrated in Figure~\ref{fig:diversity}, the performance of MCC-KD exhibits a corresponding improvement with increasing diversity among the rationales, as observed on both the ASDiv and SVAMP development sets.

\begin{figure}[t]
    \centering
    \begin{adjustbox}{width=0.45\textwidth}
        \includegraphics{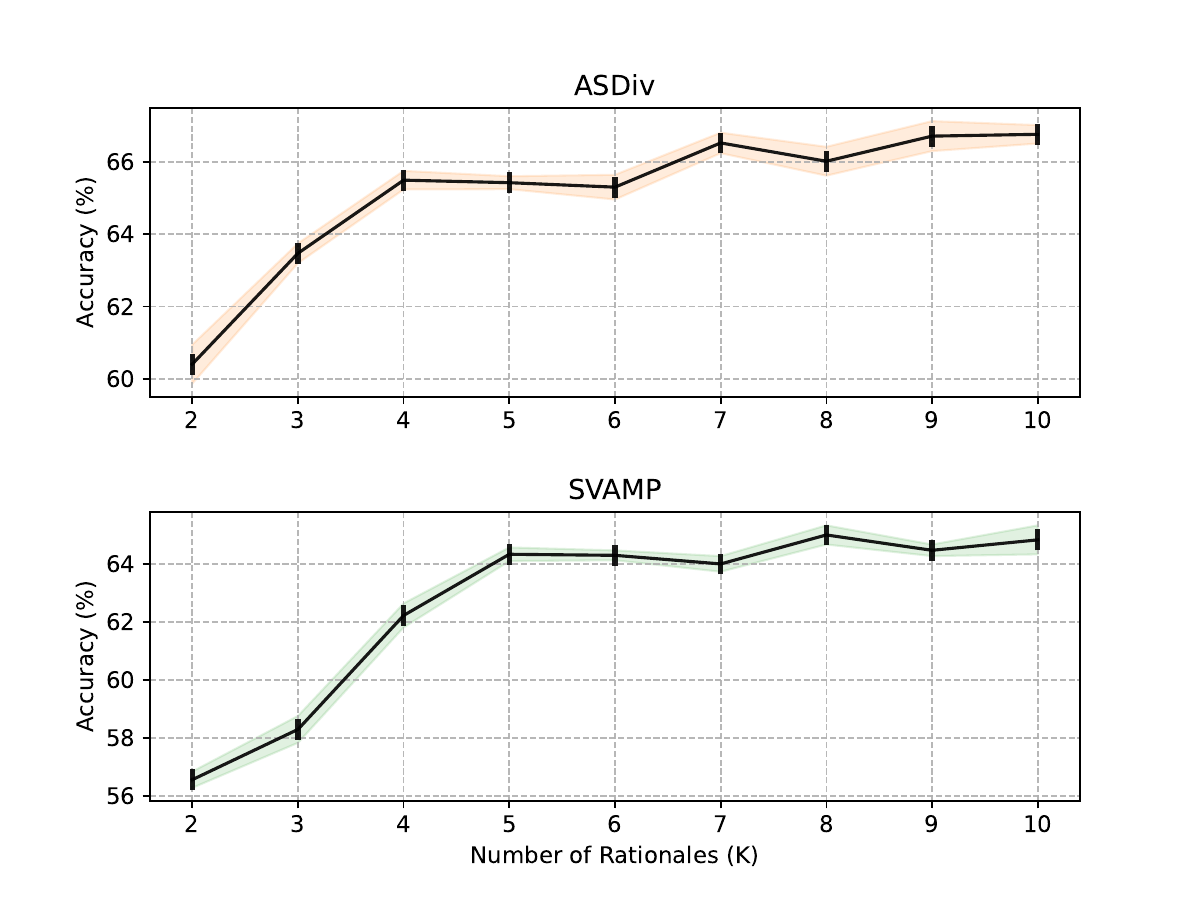}
    \end{adjustbox}
    \caption{The performance of MCC-KD on ASDiv and SVAMP with different numbers of rationales. }
    \label{fig:quantity}
    \vspace{-3mm}
\end{figure}

\subsection{The Number of Rationales}
We also examine the performance of MCC-KD with varying numbers of rationales on the SVAMP and ASDiv datasets. Note that for each training epoch, our method randomly selects two distinct rationales from a set of $K$ rationales per question as the training instances. Hence, we modify the value of $K$ for each question to assess its impact. To ensure sufficient training, we empirically set the number of training epochs to 24. The student model employed in these experiments is LLaMA-7B. As depicted in Figure~\ref{fig:quantity}, we observe that as the number of rationales increases, the model's performance on both the ASDiv and SVAMP datasets improves correspondingly. Specifically, when the number of rationales is increased from 2 to 5, there is a significant enhancement in performance on both datasets. However, when the number is further increased from 5 to 10, the performance gains become less pronounced. Therefore, taking into account computational efficiency, we opt to use 5 rationales in our experiments.

\begin{figure}
    \centering
    \vspace{-0mm}
    \begin{adjustbox}{width=0.45\textwidth}
        \includegraphics{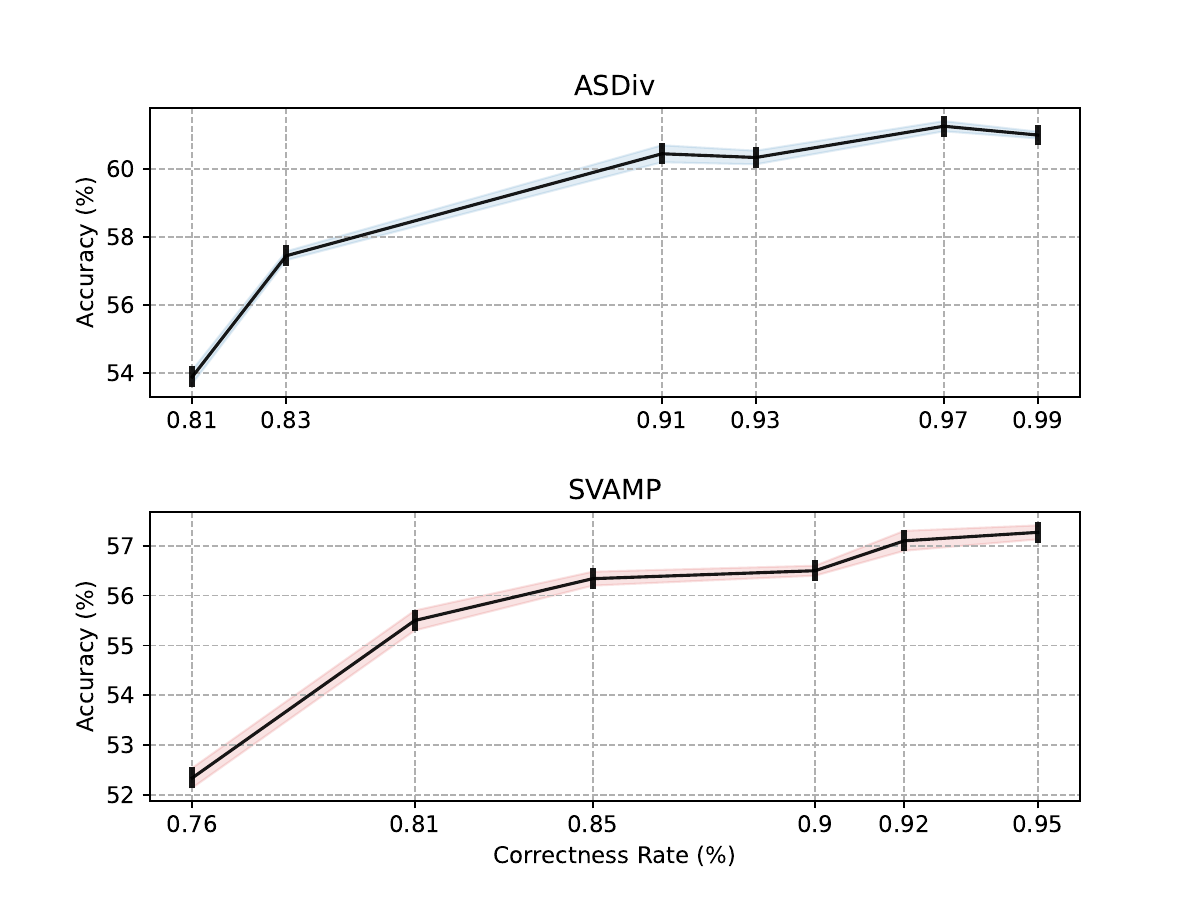}
    \end{adjustbox}
    \caption{The performances of MCC-KD on the ASDiv and SVAMP datasets with different correctness rates of the teacher generated rationales.}
    \label{fig:correctness}
    \vspace{-0mm}
\end{figure}

\subsection{Rationale Correctness}
Ensuring LLMs to generate completely accurate rationales poses a challenge, especially when dealing with complex reasoning datasets. Striking a balance between a higher correctness rate and larger data quantity is crucial, considering the increased expenses associated with LLMs' API calls. To investigate the impact of the correctness rate for teacher-generated rationales, we randomly select two rationales from the raw rationales as training instances for each question. 
% The correctness of each rationale is evaluated by comparing its predicted answer with the ground truth label answer.
We approximate the correctness of rationales by comparing the predicted answers with the ground-truth answers.
% we use the comparison between predicted answers and ground-truth answers as an approximation of the correctness of rationales.
Figure \ref{fig:correctness} illustrates the impact of the correctness rate of the generated rationales on the ASDiv and SVAMP datasets, using LLaMA-7B as the student model. As depicted in the figure, there is minimal performance difference when the correctness rate exceeds 90\%. However, a significant degradation in model performance is observed when the correctness rate falls around 80\% or below. To maintain high performance, we ensure a correctness rate of over 90\% throughout our experiments.

\subsection{Consistency Weight}
\begin{figure}
    \centering
    \begin{adjustbox}{width=0.45\textwidth}
        \includegraphics{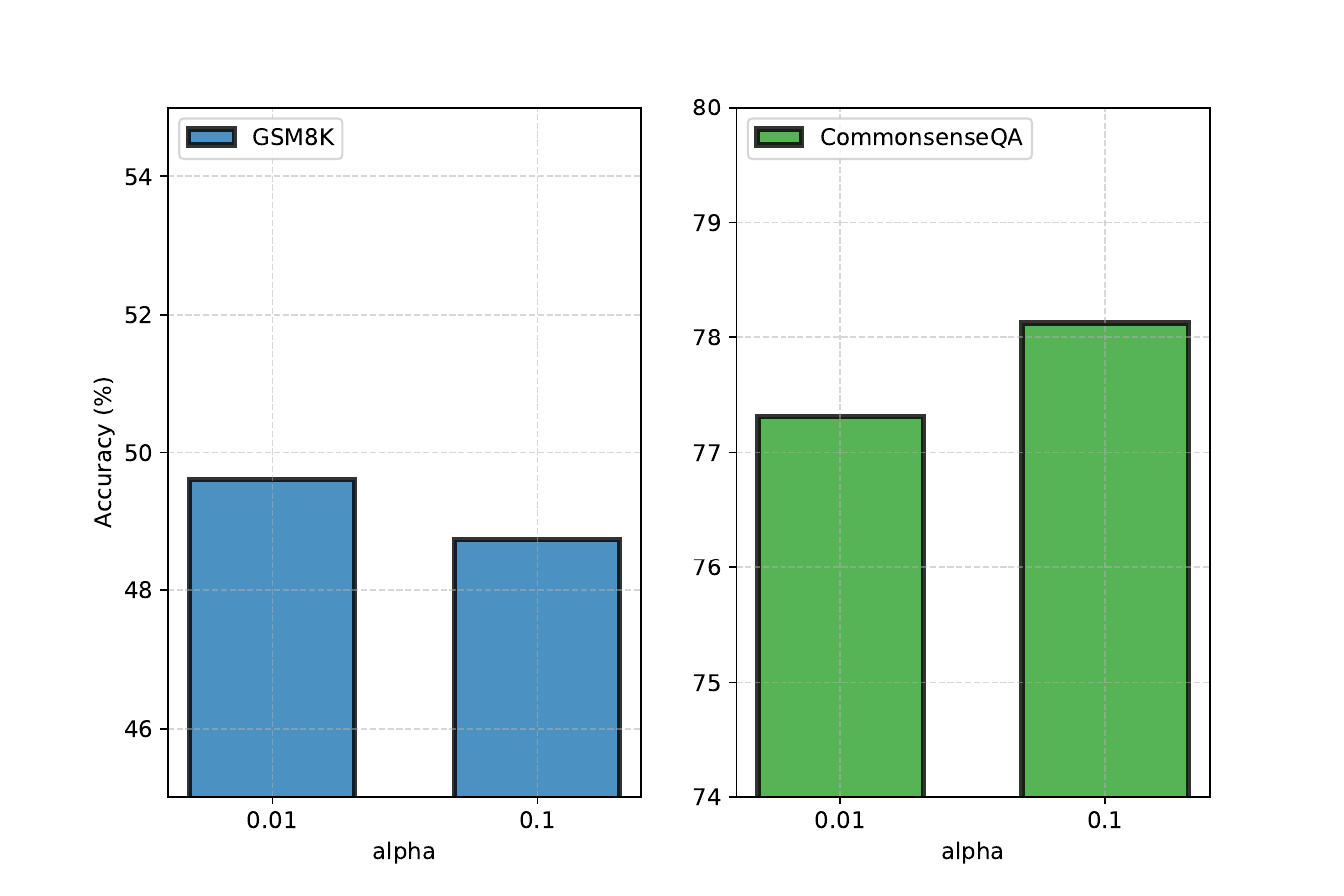}
    \end{adjustbox}
    \caption{Results on GSM8K and CommonsenseQA development sets with different consistency weights ($\alpha$).}
    \label{fig:alpha}
    \vspace{-0mm}
\end{figure}

In the training objective of MCC-KD, we introduce hyperparameter $\alpha$ to balance the multi-CoT consistency constraint. To investigate its impact on the performance of LLaMA-13B, we present results for two different values of $\alpha$ on the GSM8K and CommonsenseQA datasets in Figure~\ref{fig:alpha}. The results indicate that our method shows a preference for a smaller $\alpha$ of 0.01 on the mathematical reasoning task of GSM8K, while it favors a larger $\alpha$ of 0.1 on the commonsense reasoning task of CommonsenseQA. Therefore, we empirically select $\alpha$ values close to 0.01 for mathematical reasoning datasets (GSM8K, ASDiv, and SVAMP), and close to 0.1 for the commonsense reasoning dataset (CommonsenseQA) throughout our experiments.

\subsection{Combining with Self-Consistency}
We note that self-consistency \citep{self-consistency} also employs a consistency strategy for maintaining consistency among diverse rationales, which shares similarities with our method. However, self-consistency requires LLMs to generate multiple rationales and answers during the inference phase, determining the final answer based on the highest vote count. In contrast, our MCC-KD involves the imposition of consistency constraints on diverse rationales during the training phase. 

In this section, we investigate the incorporation of self-consistency (SC, 5 rationales for voting) and MCC-KD on the GSM8K dataset, employing the LLaMA-7B model. As observed in Table \ref{tab:self-consistent}, while both MCC-KD and self-consistency can improve the Vanilla KD approach, the performance of MCC-KD can be further enhanced through the application of self-consistency. Note that MCC-KD and self-consistency work in distinct phases, making direct comparisons of their results inappropriate. 

\begin{table}[ht]
    \centering
    \renewcommand{\arraystretch}{1.2}
    \begin{adjustbox}{width=0.48\textwidth}
        \begin{tabular}{l|c|c}
        \hline
        Models & \#Params & GSM8K \\
        \hline
        \hline
        LLaMA-7B (Vanilla KD) & 7B & 38.01 \\
        LLaMA-7B (Vanilla KD + SC) & 7B & 40.06 \\
        LLaMA-7B (MCC-KD) & 7B & 41.58 \\
        LLaMA-7B (MCC-KD + SC) & 7B & 42.49 \\
        \hline
        \end{tabular}
    \end{adjustbox}
    \caption{Results of combining our MCC-KD approach and the self-consistency (SC) method.}
    \label{tab:self-consistent}
\end{table}

\section{Conclusion}

In this paper, we propose Multi-CoT Consistent Knowledge Distillation (MCC-KD) to transfer reasoning capabilities from larger language models (LLMs) to smaller models. The primary objective is to address the diversity and consistency challenges present in existing knowledge distillation methods for this purpose. Our approach leverages multiple rationales for each given question and focuses on improving their consistency in predicting the answer. Extensive experiments are conducted using different model architectures, including LLaMA and FlanT5, and a range of model scales, such as 3B, 7B, 11B, and 13B. The experiments cover both mathematical and commonsense reasoning benchmarks. The results clearly demonstrate the superior performance of MCC-KD on both in-distribution and out-of-distribution tasks. These findings confirm that MCC-KD enhances the stability and generalizability of the student models.

\section*{Acknowledgement}

This work was supported by the National Natural Science Foundation of China (No. 62176270) and the Guangdong Basic and Applied Basic Research Foundation (No. 2023A1515012832).

\section*{Limitations}

There are three potential limitations of our work. First, the reliance on LLMs for generating rationales introduces a potential limitation in terms of cost associated with API calls. Second, there still exists a significant gap between the student model and the teacher model in mathematical reasoning tasks, requiring future efforts to reduce this disparity.
Third, this work focuses solely on exploring only one single teacher model, overlooking the potential benefits and insights that could arise from considering different LLMs as the teachers.

\bibliography{anthology,custom}
\bibliographystyle{acl_natbib}

\clearpage

\appendix

\section{Models}
\label{sec:model}
\subsection{Model Configurations}

We employ four distinct models, namely FlanT5-XL, FlanT5-XXL, LLaMA-7B, and LLaMA-13B, as our student backbone models. The FlanT5 models utilize an encoder-decoder Transformer architecture, whereas the LLaMA models adopt a decoder-only Transformer architecture. Note that the FlanT5 models have undergone instruction tuning, whereas the LLaMA models have not. Based on our empirical observations, the FlanT5 models exhibit stronger performance in commonsense reasoning tasks, while the LLaMA models excel in mathematical reasoning tasks. The configurations of these models are provided in Table \ref{tab:appendix-model-config}.

\begin{table}[ht]
    \centering
    \begin{adjustbox}{width=0.48\textwidth}
        \begin{tabular}{l|ccccc}
        \hline
        Models & \makecell[c]{Hidden \\ Size} & \makecell[c]{Attention \\ Heads} & \makecell[c]{Intermediate \\ Size} & \makecell[c]{Encoder \\ Layers} & \makecell[c]{Decoder \\ Layers} \\
        \hline
        FlanT5-XL & 2048 & 32 & 5120 & 24 & 24 \\
        FlanT5-XXL & 4096 & 64 & 10240 & 24 & 24 \\
        LLaMA-7B & 4096 & 32 & 11008 & - & 32 \\
        LLaMA-13B & 5120 & 40 & 13824 & - & 40 \\
        \hline
        \end{tabular}
    \end{adjustbox}
    
    \caption{Model configurations.}
    \label{tab:appendix-model-config}
\end{table}

\subsection{Experiment Settings}

We train all the student models on GeForce RTX 4090 GPUs using the model parallelism technique. For further accelerating the training and saving memory, we utilize quantization techniques and LoRA \citep{lora}. We apply LoRA to four weight matrices in the attention module ($W_q,W_k,W_v,W_o$) and three weight matrices in the MLP module. Similar to the objective function proposed by \citet{knowledge-distilling}, we search for the temperature parameter in the KL function. Through all experiments, we use Adam \citep{Kingma2014AdamAM} as our optimizer, and we set the learning rate to be 1e-5 in most datasets. With LLaMA-7B as the backbone model, we employ gradient accumulation, with mini-batch size of 2 and accumulation steps of 2. For different datasets, the value of $\alpha$ ranges from 0.01 to 0.1. See Table~\ref{tab:hyperparameter} for more details.

\begin{table}[ht]
    \centering
    \begin{adjustbox}{width=0.48\textwidth}
        \begin{tabular}{l|cccc}
            \hline
            Hyperparameter & GSM8K & ASDiv & SVAMP & \makecell[c]{Common \\ SenseQA} \\
            \hline
            \hline
            Learning Rate & 1e-5 & 1e-5 & 1e-5 & 1e-5 \\
            Total Batch Size & 4 & 4 & 4 & 4 \\
            Epochs & 12 & 18 & 18 & 12 \\
            $\alpha$ & 0.01 & 0.01 & 0.01 & 0.1 \\
            \hline
            \# GPUs & 4 & 4 & 4 & 4 \\
            Training Time & 24hr & 9hr & 4.5hr & 36hr \\
            \hline
        \end{tabular}
    \end{adjustbox}
    \caption{Hyperparameter settings and training cost of our method with LLaMA-7B on different datasets.}
    \label{tab:hyperparameter}
\end{table}

\subsection{Smaller Student Model}

Prior endeavors \citep{sanh2019distilbert,sun-etal-2019-patient} in the field of knowledge distillation have primarily focused on using smaller models as the recipients of knowledge transfer. The primary focus of our research resides in the transference of reasoning capabilities from LLMs to more compact counterparts, employing a chain-of-thought (COT) prompting approach. Notably, existing research \citep{chain-of-thought,fu2023specializing,kojima2022large} has demonstrated that the capacity for intricate reasoning within LLMs is typically inherent to larger models with over 100 billion parameters. 

We conduct an experiment utilizing FlanT5-base (250M) on the SVAMP and ASDiv datasets. The results presented in Table \ref{tab:smaller-results} indicate that when employing a smaller model as the student, the improvements achieved through knowledge distillation are relatively modest in comparison to those observed with larger counterparts. This observation suggests that the limited improvement attained with a smaller student model stems primarily from the model's inherent reasoning capacity rather than the effectiveness of the distillation method.

\begin{table}[ht]
    \centering
    \renewcommand{\arraystretch}{1.2}
    \begin{adjustbox}{width=0.48\textwidth}
        \begin{tabular}{l|c|cc}
        \hline
        Models & \#Params & SVAMP & ASDiv \\
        \hline
        \hline
        FlanT5-Base (Vanilla KD) & 250M & 0.0696 & 0.0687 \\
        FlanT5-Base (MCC-KD) & 250M & 0.0718 & 0.0723 \\
        \hline
        \end{tabular}
    \end{adjustbox}
    \caption{Results of FlanT5-Base on SVAMP and ASDiv.}
    \label{tab:smaller-results}
\end{table}

\section{Datasets}
\label{sec:dataset}

\subsection{Mathematical Reasoning Datasets}

For mathematical reasoning, we mainly use GSM8K \citep{cobbe2021training}, ASDiv \citep{miao-etal-2020-diverse} and SVAMP \citep{patel-etal-2021-nlp} as our training datasets. Following \citet{ho2022reasoning-teacher} and \citet{fu2023specializing}, we perform a sample-wise random split with a train-dev-test ratio of 70:15:15 (Table~\ref{tab:dataset_split}). Whereas the GSM8K dataset has the official split of training and testing sets but no development set, we randomly split the original test set with a dev-test ratio of 50:50. In order to keep consistency with the GSM8K dataset and SVAMP dataset, we only evaluate the models' mathematical reasoning abilities on the ASDiv dataset by filtering out samples with non-numeric answers. Since the MultiArith \citep{roy-roth-2015-solving}, SingleEq \citep{koncel-kedziorski-etal-2015-parsing} and AddSub \citep{hosseini-etal-2014-learning} datasets are too small, we use these datasets for our out-of-distribution experiments only (Table~\ref{tab:dataset-ood}). We don't train on the AQuA \citep{ling-etal-2017-program} dataset, as it has 100,000 examples, which is too costly to infer with an LLM.

\begin{table}[ht]
    \centering
    \renewcommand{\arraystretch}{1.2}
    \begin{adjustbox}{width=0.48\textwidth}
        \begin{tabular}{l|ccc}
        \hline
        Datasets & Train Size & Dev Size & Test Size \\
        \hline
        \hline
        GSM8K & 7473 & 660 & 659 \\
        ASDiv & 1462 & 313 & 314 \\
        SVAMP & 700 & 150 & 150 \\
        CommonsenseQA & 8520 & 1221 & 1221 \\
        \hline
        \end{tabular}
    \end{adjustbox}
    \caption{Statistics of training, development and test sets.}
    \label{tab:dataset_split}
\end{table}

\begin{table}[ht]
    \centering
    \renewcommand{\arraystretch}{1.2}
    \begin{adjustbox}{width=0.48\textwidth}
        \begin{tabular}{l|c||l|c}
        \hline
        \makecell[l]{Mathematical \\ Datasets} & Dataset Size & \makecell[l]{Commonsense \\ Datasets} & Dataset Size \\
        \hline
        \hline
        MultiArith & 600 & StrategyQA & 2290 \\
        AddSub & 395 & Date Understanding & 369 \\
        SingleEq & 508 & Shuffled Objects & 750 \\
          &   & Coin Filp & 500 \\
        \hline
        \end{tabular}
    \end{adjustbox}
    
    \caption{Datasets used to evaluate model's out-of-distribution generalization ability and the dataset sizes.}
    \label{tab:dataset-ood}
\end{table}

% =========================================

\begin{table*}[ht]
    \centering
    \vspace{-0mm}
    \renewcommand{\arraystretch}{1.2}
    \begin{adjustbox}{width=0.95\textwidth}
    \begin{tabularx}{\textwidth}{c|X}
         \hline
         \textbf{Question} & Is Tange Sazen hypothetically an ideal choice for a secretary job? Answer Choices: A. Yes B. No \\
         \hline
         \textbf{Response} & Sorry, as an AI language model, I cannot provide a definite answer as to whether Tange Sazen would be an ideal choice for a secretary job as I do not have sufficient knowledge on that individual. However, if provided with their relevant details, I could list out their qualities that may make them suitable for a secretary position. \\
         \hline
         \hline
         \textbf{Question} & Do Bing (search engine) searches earn the searcher more than competitors do? Answer Choices: A. Yes B. No \\
         \hline
         \textbf{Response} & As an AI language model, I do not have access to current market data or the ability to keep information about such kind of advertisement policies which exactly pay if someone searches. \\
         \hline
    \end{tabularx}
    \end{adjustbox}
    \caption{Examples of questions that GPT-3.5-Turbo refuses to answer from the StrategyQA dataset.}
    \label{tab:refuse-strategyQA}
    \vspace{5mm}
\end{table*}

\begin{table*}[ht]
    \vspace{-0mm}
    \centering
    \renewcommand{\arraystretch}{1.1}
    \begin{adjustbox}{width=0.95\textwidth}
    \begin{tabularx}{\textwidth}{c|X}
         \hline
         \textbf{Question} & Where do you put your grapes just before checking out? Answer Choices: A. mouth B. grocery cart C. super market D. fruit basket E. fruit market \\
         \hline
         \textbf{Response} & When grocery shopping and you find yourself with grapes you want to purchase, it is customary to put them in your grocery cart. The cart is usually provided by the store as a means to transport your selected items. Some stores may also offer fruit baskets or bags specifically made for holding grapes to prevent them from rolling around or getting squished. In the case of a fruit market, you may place your grapes in a basket provided by the vendors. As an AI language model, I do not have a personal choice, but I can provide information based on facts. \\
         \hline
         \hline
         \textbf{Question} & Before getting a divorce, what did the wife feel who was doing all the work? Answer Choices: A. harder B. anguish C. bitterness D. tears E. sadness \\
         \hline
         \textbf{Response} & It is difficult to determine the exact emotions felt by the wife who was doing all the work before getting a divorce as everyone's experiences and emotions are unique. However, she may have felt overwhelmed and frustrated by her spouse's perceived lack of effort or help in shouldering the household responsibilities. She may have also felt resentment and bitterness towards her spouse for not sharing the workload and contributing equally. Ultimately, these feelings could have led to sadness, anguish, or tears as the relationship deteriorated and ended in a divorce. As an AI language model, I do not have personal choices or emotions. \\
         \hline
    \end{tabularx}
    \end{adjustbox}
    
    \caption{Examples of questions that GPT-3.5-Turbo refuses to answer from the CommonsenseQA dataset.}
    \label{tab:refuse-commonsenseQA}
    
\end{table*}

\subsection{Commonsense Reasoning Datasets}

For commonsense reasoning, we select CommonsenseQA \citep{talmor-etal-2019-commonsenseqa} as our training dataset. 
% For the CommonsenseQA dataset split, due to the large size of the original training set, we randomly split total 1221 of instances from training set, the same size of the original development set, as our test set, 
We perform a random split of 1221 instances from the original training set to create the test set. This number corresponds to the size of the CommonsenseQA's original development set (Table~\ref{tab:dataset_split}). We do not use the Date Understanding, Tracking Shuffled Objects and Coin Flip as our training datasets due to the small size of those datasets (Table~\ref{tab:dataset-ood}), but we use them to evaluate model's out-of-distribution performance. The strategyQA \citep{geva-etal-2021-aristotle} dataset may contain private or personal information which GPT-3.5-Turbo refuses to answer (Table \ref{tab:refuse-strategyQA}). For the CommonsenseQA datasets, the GPT-3.5-Turbo may also refuse to answer certain questions (Table~\ref{tab:refuse-commonsenseQA}). However, the proportion of such cases is much smaller compared with StrategyQA, and it still provides useful information in the rationales even when the teacher model refuses to give a choice. % while the rationales in StrategyQA dataset seldom have.

For all commonsense reasoning datasets, we utilize choices-answer format for each question. The CommonsenseQA dataset has five choices per question, and the Tracking Shuffled Objects has three choices per question. See Table~\ref{tab:dataset-shuffled} for more details. There are no choices for the answers in the CoinFlip and the StrategyQA datasets. Since the answers in CoinFlip and StrategyQA are in the form of ``Yes'' or ``No'',  we transform them into binary choices-answer format, for example: (A) Yes and (B) No. Examples of the transformed datasets can be seen in  Table~\ref{tab:refuse-strategyQA} and Table~\ref{tab:dataset-coin}. 

% As for every question in the choices-answer format, we concatenate the label choice and its corresponding contents together as the answer, for example: A. jewelry store.

\begin{table*}[ht]
    \centering
    \vspace{-5mm}
    \renewcommand{\arraystretch}{1.2}
    \begin{adjustbox}{width=0.95\textwidth}
    \begin{tabularx}{\textwidth}{c|X}
         \hline
         \textbf{Question} & Alice, Bob, and Claire are dancers at a square dance. At the start of a song, they each have a partner: Alice is dancing with Sam, Bob is dancing with Helga, and Claire is dancing with Karl. Throughout the song, the dancers often trade partners. First, Claire and Alice switch partners. Then, Bob and Alice switch partners. Finally, Claire and Bob switch partners. At the end of the dance, Alice is dancing with Which choice is true ? Answer Choices: A. Sam. B. Helga. C. Karl. \\
         \hline
         \textbf{Label} & B. Helga. \\
         \hline
         \hline
         \textbf{Question} & Alice, Bob, and Claire are playing a game. At the start of the game, they are each holding a ball: Alice has a brown ball, Bob has a red ball, and Claire has a purple ball. As the game progresses, pairs of players trade balls. First, Bob and Claire swap balls. Then, Alice and Claire swap balls. Finally, Alice and Bob swap balls. At the end of the game, Bob has the Which choice is true ? Answer Choices: A. brown ball. B. red ball. C. purple ball. \\
         \hline
         \textbf{Label} & B. red ball. \\
         \hline
         \hline
         \textbf{Question} & Alice, Bob, and Claire are holding a white elephant gift exchange. At the start of the event, they are each holding a present of a different color: Alice has a blue present, Bob has a brown present, and Claire has a white present. As the event progresses, pairs of people swap gifts. First, Claire and Alice swap their gifts. Then, Bob and Claire swap their gifts. Finally, Alice and Bob swap their gifts. At the end of the event, Claire has the. Which choice is true ? Answer Choices: A. blue present. B. brown present. C. white present. \\
         \hline
         \textbf{Label} & B. brown present. \\
         \hline
    \end{tabularx}
    \end{adjustbox}
    \caption{Examples from the Shuffled Objects dataset.}
    \label{tab:dataset-shuffled}
\end{table*}

\begin{table*}[ht]
    \vspace{-0mm}
    \centering
    \renewcommand{\arraystretch}{1}
    \begin{adjustbox}{width=0.95\textwidth}
    \begin{tabularx}{\textwidth}{c|X}
         \hline
         \textbf{Question} & A coin is heads up. Jeff does not flip the coin. Jen flips the coin. Giselle flips the coin. Noel does not flip the coin. Is the coin still heads up? Note that "flip" here means "reverse". Answer Choices: A. Yes B. No \\
         \hline
         \textbf{Label} & A. Yes \\
         \hline
         \hline
         \textbf{Question} & A coin is heads up. Rena does not flip the coin. Devon does not flip the coin. Rosalinda does not flip the coin. Paulina does not flip the coin. Is the coin still heads up? Note that "flip" here means "reverse". Answer Choices: A. Yes B. No \\
         \hline
         \textbf{Label} & A. Yes \\
         \hline
         \hline
         \textbf{Question} & A coin is heads up. Dorian flips the coin. Mayra flips the coin. Freddie does not flip the coin. Magaly flips the coin. Is the coin still heads up? Note that "flip" here means "reverse". Answer Choices: A. Yes B. No \\
         \hline
         \textbf{Label} & B. No \\
         \hline
    \end{tabularx}
    \end{adjustbox}
    \caption{Examples from the Coin Flip dataset.}
    \label{tab:dataset-coin}
    \vspace{-1mm}
\end{table*}

% =========================================

\subsection{Diversity of Generated Rationales}
\label{sec:appendix-diversity}
When the sampling temperature is too low (e.g., $\tau=1$), the teacher model tends to generate the same rationales for the same question repeatedly. On the other hand, when the temperature is too high (e.g., $\tau=1.5$), it can lead to a significant decrease in the quality of the generated rationales by the teacher model. Hence, we choose an appropriate temperature $\tau=1.3$ to balance the generation of diverse rationales while maintaining their quality. However, we observe that even with sampling temperature $\tau=1.3$, the teacher model may still exhibit a tendency to generate somewhat similar rationales occasionally, as shown in Table~\ref{tab:repetitiveness}. We argue that the diversity among teacher-generated rationales is a crucial factor, as higher diversity implies more diverse of solutions. We believe that there should be minimal diversity between rationales in order for effective distillation.

\begin{table*}[ht]
    \centering
    \renewcommand{\arraystretch}{1.2}
    \begin{adjustbox}{width=0.95\textwidth}
    \begin{tabularx}{\textwidth}{c|X}
         \hline
         \textbf{Question} & Seal releases 3 songs every month. He gets \$2000 per song. How much money does he get in his first 3 years? \\
         \hline
         \textbf{Rationale \#1} & He releases 12*3=36 songs a year. That means he is paid 36*2000=\$72,000 a year. So over 3 years, he will make 72,000*3=\$216,000. Therefore, the answer is 216000. \\
         \hline
         \textbf{Rationale \#2} & He releases 3*12=36 songs a year. So he makes 36*2000=\$72,000 per year. That means he earns 72000*3=\$216,000. Therefore, the answer is 216000.\\
         \hline
         \textbf{Rationale \#3} & He releases 3*12=36 songs in a year. So he makes 36*2000=\$72,000 per year. That means he earned 72,000*3=\$216,000. Therefore, the answer is 216000.\\
         \hline
         \hline
         \textbf{Question} & Ed and Jacob already had 2 shells in their shell collection. Then they went to the beach to collect even more. Ed found 7 limpet shells, 2 oyster shells, and 4 conch shells. Jacob found 2 more shells than Ed did. How many shells do the pair have altogether? \\
         \hline
         \textbf{Rationale \#1} & Ed found 7 + 2 + 4 = 13 shells. Jacob found 13 + 2 = 15 shells. In total, they collected 2 + 13 + 15 = 30 shells. Therefore, the answer is 30. \\
         \hline
         \textbf{Rationale \#2} & Ed found a total of 7 + 2 + 4 = 13 shells. Jacob found 13 + 2 = 15 shells. Together they found 2 + 13 + 15 = 30 shells. Therefore, the answer is 30.\\
         \hline
         \textbf{Rationale \#3} & Ed found a total of 7 + 2 + 4 = 13 shells. Jacob found 13 + 2 = 15 shells. Altogether, Ed and Jacob have 2 + 13 + 15 = 30 shells. Therefore, the answer is 30.\\
         \hline
    \end{tabularx}
    \end{adjustbox}
    \caption{Examples from the GSM8K dataset showing that despite using a high sampling temperature $\tau=1.3$, the rationales generated by GPT-3.5-Turbo still exhibit a considerable degree of similarity. }
    \label{tab:repetitiveness}
\end{table*}

\end{document}